\documentclass[Afour,sageh,times]{sagej}

\usepackage{moreverb,url}

\usepackage[colorlinks,bookmarksopen,bookmarksnumbered,citecolor=red,urlcolor=red]{hyperref}
\usepackage{algorithm}
\usepackage{algpseudocode}
\usepackage{amsmath}
\usepackage{caption}
\usepackage{multirow}
\usepackage{xcolor}
\usepackage{tcolorbox}
\usepackage{amssymb}
\usepackage{multirow}
\usepackage{subcaption}
\usepackage{todonotes}

\newcommand\BibTeX{{\rmfamily B\kern-.05em \textsc{i\kern-.025em b}\kern-.08em
T\kern-.1667em\lower.7ex\hbox{E}\kern-.125emX}}

\begin{document}

\runninghead{Tilwani et al.}

\title{NeuroSymbolic AI for Legal AI-TRISM: Trustworthy, Reliable, Interpretable, Safe Models}

\author{Deepa Tilwani\affilnum{1}, Yash Saxena\affilnum{2}, Ankur Padia\affilnum{2}, Srinivasan Parthasarathy\affilnum{3} and Manas Gaur\affilnum{2}}

\affiliation{\affilnum{1}Department of Computer Science, AI Institute, University of South Carolina, Columbia, South Carolina, USA \\
\affilnum{2}Department of Computer Science and Electrical Engineering, University of Maryland, Baltimore County, Baltimore, Maryland, USA \\ 
\affilnum{3} Department of Computer Science and Engineering, The Ohio State University, Columbus, Ohio, USA}

\corrauth{Deepa Tilwani}

\email{dtilwani@mailbox.sc.edu}

\begin{abstract}
Large Language Models (LLMs) have transformed natural language processing, but their lack of interpretable reasoning and tendency to hallucinate pose significant challenges for legal applications. While LLMs show promise for legal text analysis and generation, they struggle with accurate citation attribution and precedent verification. For example, in legal contexts, a single incorrect precedent can jeopardize a case. Current approaches to improve LLM reliability in legal domains suffer from two key limitations: inadequate integration of structured legal knowledge during training or fine-tuning, and insufficient verification mechanisms for generated legal content.
To address these challenges, we propose the TRISM (Trustworthy, Reliable, Interpretable, Safe Models) framework, which integrates NeuroSymbolic AI principles with LLMs to leverage both neural learning capabilities and symbolic reasoning over structured legal knowledge. The TRISM approach addresses the above limitations while maintaining interpretable decision pathways. Our framework formalizes the extraction of symbolic knowledge from legal textual documents and incorporates Retrieval-Augmented Generation (RAG) as a core component for grounding LLM outputs in verified legal sources. In this position paper, we make the following contributions: (1) An analysis of the limitations of AI in law; (2) Introduce RASOR RAG which creates foundations for neurosymbolic RAG by generating explicit interpretable rationales that could be formalized into symbolic representations; (3) A formalized methodology for creating symbolic legal knowledge bases that support both interpretable reasoning and output verification in LLMs; and (4) The TRISM framework for integrating symbolic legal knowledge with LLMs. 
\end{abstract}


\maketitle

\section{Introduction}

In 2024, a U.S. court sanctioned lawyers for submitting artificial intelligence (AI) generated briefs laced with fabricated citations—an incident that vividly illustrates the risks of unverified AI use in high-stakes legal contexts \citep{reuters2024aifine}. However, despite such cautionary examples, generative AI is rapidly transforming the legal profession, with growing applications in document analysis, legal research, and draft preparation \citep{usiaffinityImpactGenerative,economicsobservatoryGenerativeArtificial}. Large Language Models (LLMs) in particular hold considerable promise for efficiently processing legal documents, extracting key information, and assisting in the creation of preliminary drafts for contracts, briefs, and memoranda \citep{qin2024exploring,constant2024path,jakub2023large}.

While LLM-based AI systems can recognize legal terminology and suggest relevant precedents, their capabilities remain limited and require careful human oversight to ensure accuracy and consistency \citep{accLegalTech}\citep{legaltechtalkHarnessingPower}. Moreover, such systems face significant challenges in downstream legal applications, particularly in understanding complex legal principles and their contextual application \citep{lai2023largelanguagemodelslaw}. To better understand the problem, consider a dispute between the landlord and the tenant over an emergency repair; an LLM in a zero-shot setting can correctly identify that tenants generally have the right to withhold rent for severe habitability issues, such as a broken heating system during winter. However, the off-the-shelf model might fail to consider specific state laws and side information that require tenants to notify the landlord in writing and provide a reasonable repair period before withholding rent. Furthermore, while suggesting relevant cases about rent withholding, the LLM might miss critical local housing court decisions that require tenants to deposit the withheld rent into an escrow account. 

The above examples illustrate how LLMs can simulate understanding of basic legal principles but struggle with the nuanced requirements and procedural steps based on jurisdiction, the knowledge that experienced lawyers use daily to advise clients appropriately. 

%\todo{nice paragraph} 
Legal reasoning often requires synthesizing multiple sources of law, including constitutions, statutes, regulations, case law, and persuasive secondary materials, across overlapping jurisdictions and time frames. LLMs fail to recognize priorities between local rules and general principles, where local regulations might override general principles \citep{qin2024exploringnexuslargelanguage}. Such limitations in understanding legal hierarchies and jurisdictional nuances highlight current capabilities and limitations of LLM and the need for human legal expertise for accurate legal analysis and advice. 

Neurosymbolic AI represents a novel approach that integrates neural networks with symbolic
reasoning (e.g., Knowledge Graph (KG), legal precedents, guidelines) to help build trustworthy AI
systems \citep{10777891}. Neurosymbolic AI-based approach combines the pattern
recognition capabilities of neural networks with the logical inference
abilities of symbolic systems, addressing the limitations of each method when
used in isolation \citep{sheth2023neurosymbolicaiwhy}. The neural component processes raw data and learns patterns,
which it does very well. A symbolic element is necessary to identify both explicit and implicit entities and their relationships within the data, enabling logical reasoning and rule-based processing.
By combining symbolic components, a neural network can better follow expert instructions, understand structured knowledge (like legal guidelines or precedents), recall this knowledge when needed, and produce outputs that experts can easily understand.

Despite advances through human feedback and
training, current AI models like ChatGPT still struggle to produce information
that is both domain-safe and traceable in its reasoning \citep{bdcc5020020}. Neurosymbolic AI
addresses these limitations by incorporating structured symbolic knowledge
alongside neural patterns, leading to reduced hallucinations, improved
explainability, and enhanced reasoning capabilities. The CREST framework proposed by \cite{gaur2023buildingtrustworthyneurosymbolicai} demonstrates how neurosymbolic methods can
ensure consistency, reliability, and safety in critical applications. 

In this position paper, we make the following contributions: (1) An analysis of existing AI-driven methods and their limitations in legal applications; (2) RASOR demonstrates that transparent, structured reasoning not only improves explainability but actually delivers superior performance—cutting hallucination rates from 75\% to under 40\%. This empirical validation creates a stepping stone toward full neurosymbolic integration, supported by our formalized knowledge graph methodology; (3) A formalized methodology for creating symbolic legal knowledge bases that support both interpretable reasoning and output verification in LLMs; and (4) The TRISM framework for integrating symbolic legal knowledge with LLMs to meet key criteria for regulated domains, including accuracy, explainability, fairness, safety, and regulatory assurance.

\section{Challenges in Current Legal AI}

LLMs are trained on massive datasets with billions of tokens and often simulate the understanding of the real world. Such capabilities have provided performance gains in summarization~\citep{Licari2023}, boundary detection~\citep{qu-meng-2024-tm}, information extraction \citep{Goebel2023}, legal judgment prediction~\citep{Jiang2023}, named entity recognition~\citep{Lee2023}, and text segmentation~\citep{Aumiller2021}. However, several challenges exist as described below:

\begin{itemize}
    \item \textbf{Complex Family Law Cases}: In child custody arrangements, AI struggles to understand the nuanced family dynamics, emotional considerations, and child well-being that require a level of understanding beyond AI's current capabilities \citep{hennesseyLimitationsFirm}. Human lawyers can factor in empathy, consequences, and lived experiences to make judgments that AI cannot replicate.
    \item \textbf{Ethical Dilemmas}: Navigating ethical dilemmas in the legal context requires a deep understanding of professional responsibilities, moral principles, and the potential consequences of one's actions. The case of ``E.F. Hutton \& Co. v. Brown'' (1986) illustrates the complexities of conflicts of interest in corporate law, highlighting the importance of rigorous conflict checks and transparent communication \citep{Jessa2024-ki}.
    \item \textbf{Causal Reasoning in Law}: Legal reasoning often involves understanding cause-and-effect relationships, which LLMs struggle to grasp. This limitation becomes evident when analyzing complex chains of events or determining liability in intricate legal cases \citep{linkedinDiscoverThousands}.
    \item \textbf{Analogical Reasoning}: While GenerativeAI can identify similar cases (e.g., LegalRAG), it struggles with the nuanced task of determining which cases are truly analogous and why. This requires a deeper understanding of legal principles and their underlying rationales, which current AI systems lack \citep{sunstein2001artificial}.
    \item \textbf{Contextual Interpretation}: AI often processes information in a fragmented manner, focusing on specific words or phrases without grasping the overall context (e.g., LegalBERT \citep{chalkidis2020legal}). This can lead to misinterpretations of legal concepts. \textit{Legal language} often requires nuanced interpretation based on context. AI may struggle to differentiate between subtle variations in meaning that are clear to human legal experts. \textit{Legal reasoning} often relies on implicit understanding and unwritten principles. AI systems may miss these crucial elements, leading to incomplete or inaccurate analyses.
    \item \textbf{Hallucinations and Temporal Mismatch}: A Stanford–Yale study challenged claims by major legal AI providers, including Thomson Reuters, Casetext, and LexisNexis, that their retrieval-augmented generation (RAG) tools had eliminated hallucinations. Despite these assurances, the research found that their products still hallucinated 33\% of the time, compared to GPT-4’s 43\% \citep {magesh2024hallucination}. 
    %\todo{How is Thomson Reuters, Casetext are related with the sentence above} Although Casetext, Thomson Reuters, and LexisNexis marketed their tools as hallucination-free, the research found that their products still hallucinated 33\% of the time, compared to the rate of GPT-4 which is 43\%. In addition, outdated legal databases and suboptimal text fragmentation methods in LegalRAG systems can compromise the precision of the response by retrieving obsolete information and fragmenting critical legal context.
\end{itemize}

In order to address the challenges above a path forward lies in developing neurosymbolic systems that combine the pattern recognition strengths of neural networks with symbolic reasoning capabilities. Such hybrid systems could better represent and manipulate legal concepts, rules, and precedents while maintaining the flexibility to handle natural language. By incorporating explicit knowledge representation and logical inference mechanisms, next-generation legal AI assistants could provide more reliable analysis, better handle edge cases, and offer clearer explanations of their reasoning.

\section{Rethinking RAG: The Case for Neurosymbolic Approaches}
Recent research has revealed concerning hallucination rates in legal AI, with LLMs producing misleading responses as high as 69\% to 88\% on specific legal queries \cite{dahl2024large}. As shown in Figure~\ref{fig:1}, both GPT-4 and SaulLM-7B \cite{colombo2024saullm} fail to correctly interpret an indemnification clause from a contract present in the Contract Understanding Atticus Dataset (CUAD) \citep{hendrycks2021cuad}. 
Notably, SaulLM-52B is built on the Mixtral-54B architecture. This architecture is a Transformer-based model enhanced with Mixture of Experts (MoE) layers to improve computational efficiency and adaptability for handling extensive contexts. The fine tuning of Mixtal-54B involved extensive legal corpora to understand legal documents and generate legal text. Despite prohibitive supervised fine-tuning, the outcome generated is incorrect.

\begin{figure}[t]
    \begin{tcolorbox}[colback=yellow!5!white,colframe=yellow!50!black,
    colbacktitle=yellow!75!black]
    \textbf{Contract Excerpt:} Section 4.2: Party A shall indemnify Party B against third-party claims arising from breach of this Agreement, except that Party A shall have no obligation to indemnify for claims arising from Party B's gross negligence or willful misconduct. \\
    \textbf{Query:} Is Party A required to indemnify Party B for all third-party claims?
    \tcblower
    \textbf{GPT-4 Response:} Yes, Party A must indemnify Party B against all third-party claims arising from breach of this Agreement.\\
    \textbf{SaulLM Response:} Yes, Section 4.2 requires Party A to indemnify Party B against all third-party claims arising from breach of the Agreement.
    \end{tcolorbox}
    \caption{\bf SaulLM \& GPT-4 lack context. Tested on July 08, 2025}  
    \label{fig:1}
\end{figure}

\begin{figure*}[t]
    \centering
        \centering
        \includegraphics[width=\textwidth]{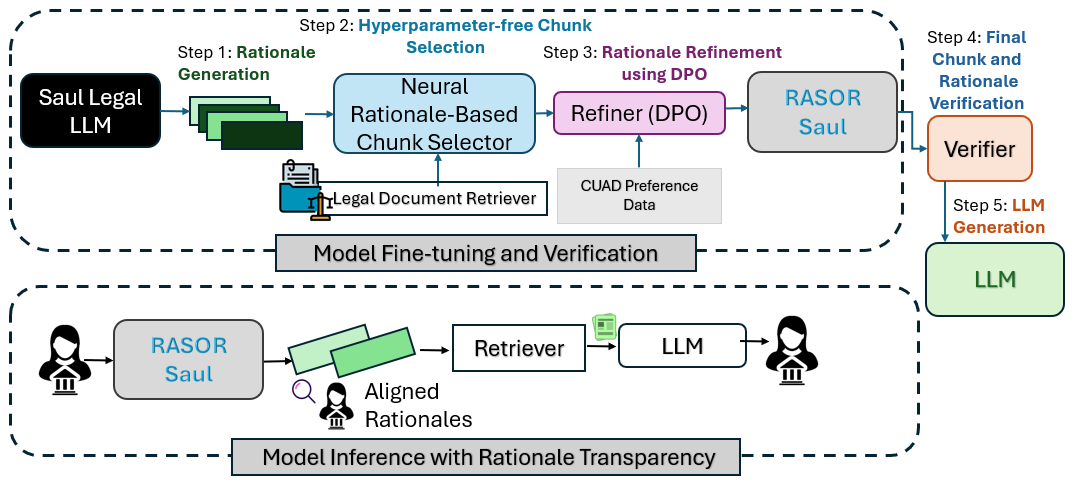}
        \caption{\bf RASOR Architecture} %Our experiment used LLAMA3-8B as LLM.}
        \label{fig:2}
\end{figure*}
Legal professionals want to know how an LLM arrives at a specific legal conclusion or recommendation. More specifically, does the model appropriately identify relevant statutes or case law? Does it appropriately weigh competing legal doctrines? As shown in Figure \ref{fig:1}, both LLMs- GPT-4 and SaulLM fail to recognize a \textit{critical exception} that limits Party A's obligations. In contrast, models focus solely on the initial requirement and overlook the qualifying context that follows. This failure highlights a fundamental limitation: these models frequently struggle to capture contextual nuance within complex legal provisions, thereby undermining their reliability in high-stakes legal interpretations.

Existing solutions, such as increasing the context window or RAG, offer some redressal but fall short of addressing the core issues of error (hallucination) and transparency \citep{barnett2024seven}. Increasing the context window increases computational complexity while generating verbose outputs that may obscure key legal reasoning. Conversely, RAG grounds responses in authoritative legal sources to help mitigate - yet recent studies report persistent hallucination rates of 17–34\% in commercial legal research tools \citep{magesh2024hallucination} . These shortcomings largely stem from RAG’s reliance on complex and often opaque design decisions, including chunk size, retrieval thresholds, and ranking heuristics, which can inadvertently distort the meaning of the retrieved content before generation, causing such errors leading to the following question: {\it How can we design an AI framework that generates legal text with clear traceability to authoritative sources with transparent, interpretable explanation?} 
To operationalize TRISM in the legal domain, we instantiate it with \textbf{RASOR}---a retrieval-and-reasoning pipeline that enforces explicit source attribution, structured rationales, and safety checks. Concretely, RASOR realizes TRISM’s pillars by (i) grounding claims in verifiable sources (\emph{Trustworthy}), (ii) stabilizing retrieval and reasoning with a modular design (\emph{Reliable}), (iii) exposing rationale steps and citation paths (\emph{Interpretable}), and (iv) applying jurisdictional and precedence constraints before finalization (\emph{Safe}). The next section details RASOR’s components and shows how this alignment translates into measurable gains on legal benchmarks (see Fig.~\ref{fig:2}).

%\todo{RASOR Architecture deserves to have its own section. Its one of the main contribution}
\section{RASOR: Implementing Neurosymbolic RAG for Legal Applications}
RASOR (\textbf{RA}tionalize, \textbf{S}elect, and \textbf{R}efine), implements TRISM’s design principles for legal text generation: it \emph{rationalizes} (interpretable reasoning), \emph{selects} (reliable retrieval), \emph{refines/verifies} (safety constraints), and \emph{structures} outputs with explicit citations (trustworthiness). This alignment guides the architecture presented below (Fig.~\ref{fig:2}): rationale generation, chunk selection, verification, and output structuring. 
%RASOR is an innovative retrieval framework that generates rationales and refinements to emphasize semantic fragment retrievals, ensuring precise answers to user queries without relying on subjective and opaque RAG heuristics. 
Through rationale generation and refinement, RASOR mitigates hallucinations and ensures transparency in chunk selection. 
Figure \ref{fig:2} %\todo{figure and reference are in different page. inconvenient to go up and down while reading the paper} 
outlines the RASOR architecture, which follows a stepwise process to generate transparent and trustworthy legal rationales. In Step 1, SaulLM is fine-tuned to generate rationales from legal datasets containing expert queries and documents. Step 2 involves matching these generated rationales with relevant document chunks using a \textit{hyperparameter-free} method called the Neural Rationale-based Chunk Selector (NRCS), which utilizes a Dense Passage Retriever (DPR). 
\begin{figure}[!htbp]
    \centering
    \begin{minipage}[t]{0.3\textwidth}
        \centering
        \includegraphics[width=\linewidth]{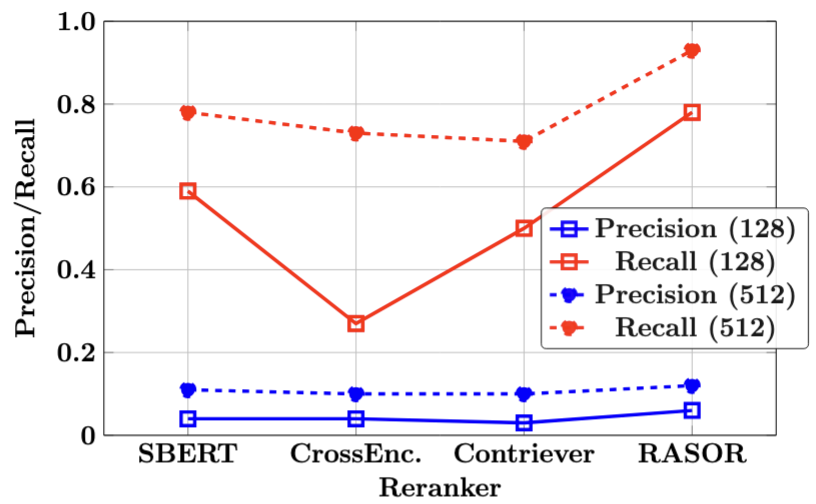}
        \subcaption{Information retrieval experiment \label{fig:comparison1}}
    \end{minipage}%
    
    \begin{minipage}[t]{0.3\textwidth}
        \centering
        \includegraphics[width=\linewidth]{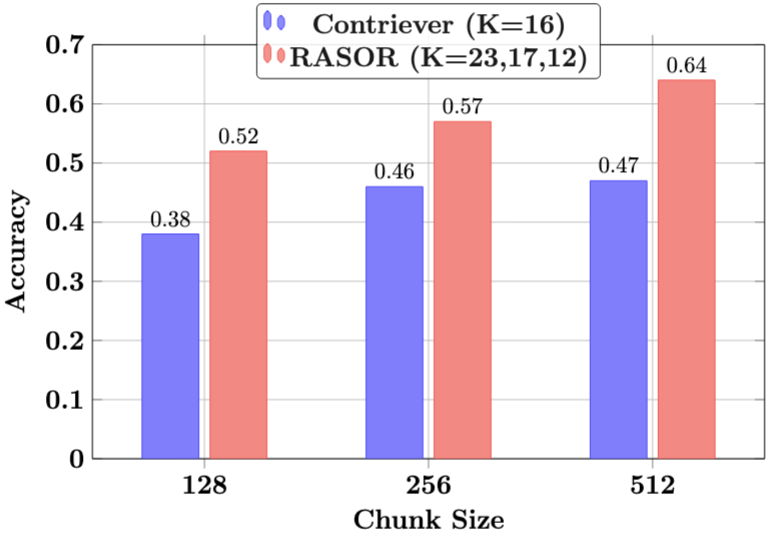}
        \subcaption{Generation accuracy computed by exact string match scoring. Generations are made by LLAMA3-8B.\label{fig:comparison2}}
    \end{minipage}
    \caption{\footnotesize RASOR's dynamic K-value selection (K=23,17,12 based on query complexity) outperforms Contriever's fixed approach (K=16), enabling LLAMA-8B to generate superior responses across all context lengths. K represents the number of top-ranked chunks.}
    \label{fig:comparisons}
\end{figure}

 In Step 3, RASOR's Refiner module applies preference learning inspired by Direct Preference Optimization (DPO) to optimize the rationales, selecting those most aligned with relevant document chunks using paired examples from the dataset CUAD. This eliminates the need for reinforcement learning models like RLHF. Step 4 introduces an unsupervised Verifier to filter out irrelevant or contradictory chunks, mitigating hallucination risks. Finally, Step 5 performs inference using verified rationales, ensuring that the output is relevant, aligned with domain knowledge, and promoting transparency and reliability.
Within the TRISM framework, RASOR is connected to a module that structures the output, organizing retrieved legal precedents and symbolic reasoning steps into a coherent message plan. This ensures that the generated response presents statutes, case law, and reasoning steps in a logical sequence, with explicit citations—directly supporting TRISM’s goals of trustworthiness, reliability, interpretability, and safety.
%\todo{what is the message plan to share? If its feasibility make it a subsection under RASOR}

\noindent\textbf{Preliminary Evaluation.}
We evaluated RASOR on the CUAD, which contains 13,000+ expert annotations across 510 contracts with an estimated \$2 million development cost, requiring 70-100 hours of training per legal annotator, making it a perfect resource to develop a responsible RAG to develop a generalizable algorithm and can be applied to proprietary data. Results were evaluated on the CUAD test set, consisting of 400 queries. 

In information retrieval evaluations on the CUAD dataset, RASOR demonstrates superior performance over all baseline re-rankers. As shown in Figure \ref{fig:comparison1}, RASOR achieves the highest precision-recall balance, with robust recall at 128 (small legal documents) and 512 (larger legal documents)-token chunks (approximately 93\%) compared to other re-rankers. Figure \ref{fig:comparison2} further highlights RASOR's effectiveness in text generation using LLAMA3-8B across different chunk sizes, where it consistently outperforms Contriever \citep{izacard2021unsupervised} (a state-of-the-art re-ranker, see Figure \ref{fig:comparison1}) by up to 17\% in accuracy. 
We compute accuracy through exact string match over similarity-based metrics, as exact string match penalizes errors (from LLM hallucination) in legal text generation. 
Such an approach ensures a strict evaluation criterion: only responses matching exact matches are scored as correct, providing an {\it unambiguous assessment of hallucination reduction}. 
At 128-token chunks, RASOR already achieves 52\% accuracy, compared to Contriever's 38\%, and extends to 64\% accuracy, compared to 47\% at 512-token chunks (long documents). 
Further, Figure \ref{fig:comparison2} shows how RASOR dynamically adjusts the k-value (top-ranked documents included) based on query complexity, optimizing retrieval while minimizing noise. RASOR achieves hallucination rates below 40\% when compared to Contriever (58-65\%) and SaulLM-52B (75\%), maintaining performance across all context lengths through dynamic K-value selection. 
RASOR consistently performs better across different context lengths, providing a robust knowledge-augmented framework for adapting large language models in the legal domain.

\subsection{RASOR as a Crucial First Step Toward Neurosymbolic Legal AI}
%RASOR serve as an essential stepping stone that propels the development of neurosymbolic AI in learning applications, based on several key transitional mechanisms. 

%RASOR shows us something important: transparent reasoning isn't just a nice-to-have feature—it actually works better. 
%The system reduces hallucination rates substantially, from approximately 75\% with SaulLM-52B to below 40\%. This shift is more than a marginal improvement—it represents the difference between a model that is incorrect in three out of four responses and one that achieves a level of reliability suitable for practical legal applications. Notably, these results demonstrate that explainable AI offers benefits beyond enhancing user confidence; it tangibly improves accuracy. When AI systems articulate their reasoning process step by step, error rates decline. This dual advantage—greater transparency and improved reliability—underscores the value of investing in advanced reasoning architectures for legal AI, rather than relying on opaque, black-box models incapable of justifying their conclusions.
RASOR demonstrates that explainable, domain-adapted AI can deliver meaningful performance gains in legal applications. In controlled evaluations on the CUAD dataset—a benchmark for nuanced legal clause detection and contract analysis—hallucination rates dropped from roughly 75\% with SaulLM-52B to under 40\% with RASOR. This improvement shifts the system from being largely unreliable to one viable for real legal work. The gain stems from RASOR’s ability to present its reasoning process step by step, which both increases user trust and directly reduces errors. By combining retrieval, symbolic reasoning, and structured output, RASOR delivers the transparency legal professionals require and the accuracy they demand—making a strong case for investment in neurosymbolic approaches over opaque, black-box systems. Within the TRISM framework, RASOR functions as a concrete instantiation of trustworthy, reliable, interpretable, and safe model design, illustrating how each pillar can be operationalized.

%RASOR's architecture creates a conceptual bridge between pure neural approaches and neurosymbolic systems by establishing a modular pipeline that can be systematically enhanced with symbolic components. Currently, RASOR follows a purely neural pathway where user queries trigger neural rationale generation, followed by neural chunk selection, neural verification, and final generation. However, this architecture naturally lends itself to evolutionary enhancement toward neurosymbolic integration. Each neural component in RASOR's pipeline has a transparent symbolic counterpart: neural rationale generation can evolve into symbolic rule application based on formal legal ontologies, neural chunk selection can be replaced with knowledge graph-based retrieval using structured legal relationships, neural verification can be upgraded to logical verification using formal constraints and precedence rules, and the final generation step can incorporate neurosymbolic reasoning that combines neural language capabilities with symbolic logical inference. 
\textbf{Architecture and Evolutionary Potential:} RASOR's architecture creates a conceptual bridge between pure neural approaches and neurosymbolic systems via a modular pipeline that can be systematically enhanced with symbolic components. Currently, RASOR follows a purely neural pathway where user queries trigger neural rationale generation, followed by neural chunk selection, neural verification, and final generation. However, each neural component has a transparent symbolic counterpart: neural rationale generation can evolve into symbolic rule application over formal legal ontologies; neural chunk selection can be replaced with knowledge graph-based retrieval using structured legal relationships; neural verification can be upgraded to logical verification with formal constraints and precedence rules; and final generation can incorporate neurosymbolic reasoning that combines neural language capabilities with symbolic inference. 
%RASOR's notable performance on the CUAD dataset reveals three important insights that build upon each other. First, when we tested structured reasoning against traditional black-box approaches in legal contexts, the structured approach consistently outperformed, proving that legal AI benefits from systems that can demonstrate their work rather than just producing answers without explanation. Second, and perhaps more importantly, legal professionals embraced these transparent systems once they saw the better results, which breaks down a major barrier to adoption that many AI researchers worry about. Third, the fact that RASOR was explicitly designed for legal tasks rather than being a general-purpose tool made all the difference—domain expertise matters in this field. These three findings, together, create something valuable: a proven foundation that future researchers can build upon. Instead of starting from scratch with theoretical arguments about why neurosymbolic approaches might work in legal AI, we now have concrete evidence that structured, transparent, domain-specific systems deliver real improvements that legal professionals will use. This gives researchers the confidence and justification they need to invest in even more sophisticated neurosymbolic architectures, knowing there's already a track record of success to build upon.

Our evaluation revealed three key insights. First, structured reasoning consistently outperformed black-box approaches, proving that legal AI benefits from systems that can ``show their work” instead of producing unexplained answers. Second, legal professionals embraced these transparent systems once they saw the improved outcomes—overcoming a major barrier to adoption often cited by AI researchers. Third, RASOR’s design with specialized legal expertise was crucial: domain expertise amplified performance gains that generic LLMs could not match. Together, these findings form a proven foundation for future research—moving from theoretical arguments about neurosymbolic potential in law to concrete evidence that structured, transparent, domain-specific systems deliver real, adoptable improvements.

RASOR's modular design creates natural extension points for symbolic integration, with LLM-era implementation pathways for each component: The \textit{NRCS} module can be replaced by a KG-enhanced chunk selector using legal ontologies. This can be implemented by building legal knowledge graphs using LLM-based construction methods such as GraphRAG \citep{edge2024local} and pairing it with G-Retriever \citep{he2024gretriever} for graph-aware retrieval. In addition, instruction-tuned retrievers \citep{asai2023task} can allow RAG systems to adapt retrieval strategies based on legal query types dynamically. In contrast, Legal-BERT legal domain embeddings can enhance semantic understanding of legal concepts within the knowledge graph structure \citep{chalkidis2020legal}. The \textit{unsupervised verifier} responsible for filtering contradictory chunks can be replaced by a Rule-based Logical Verifier using formal legal constraints. This can be implemented by replacing the unsupervised verifier with LLM-based logical reasoning using chain-of-thought methodologies \citep{wei2022chain} enhanced with tool-augmented reasoning \citep{schick2023toolformer}. Implement formal legal constraint checking using constitutional AI principles \citep{bai2022constitutional} to ensure generated rationales comply with legal precedence hierarchies. Multistep verification can be achieved through self-consistency decoding \citep{wang2022self}, where multiple reasoning paths validate legal conclusions against formal constraints. The DPO-based preference optimization can be replaced by symbolic reasoning using legal precedence hierarchies. This can be achieved by enhancing the DPO framework with legal precedence-aware optimization, utilizing hierarchical preference learning \citep{rafailov2024direct} \citep{saxena2025ranking}. 
Implement multi-agent legal reasoning systems \citep{wu2023autogen} where specialized agents handle different legal domains (constitutional law, statutory interpretation, case precedents) and coordinate through structured legal reasoning protocols. Citation network analysis \citep{dadgostari2021modeling} can inform precedence weighting, while constitutional AI approaches \citep{bai2022constitutional} ensure adherence to legal hierarchy principles.
Each extension leverages modern LLM capabilities while maintaining compatibility with established legal reasoning patterns, creating a progressive pathway from neural rationalization to complete neurosymbolic legal reasoning systems.

\section{From Implementation to Architecture: Neurosymbolic Legal AI Systems}
%\todo[inline]{Deepa, Yash - Draw a pipeline diagram describing neurosymbolic RAG}

\begin{figure}[t]
    \centering
    \includegraphics[width=\linewidth]{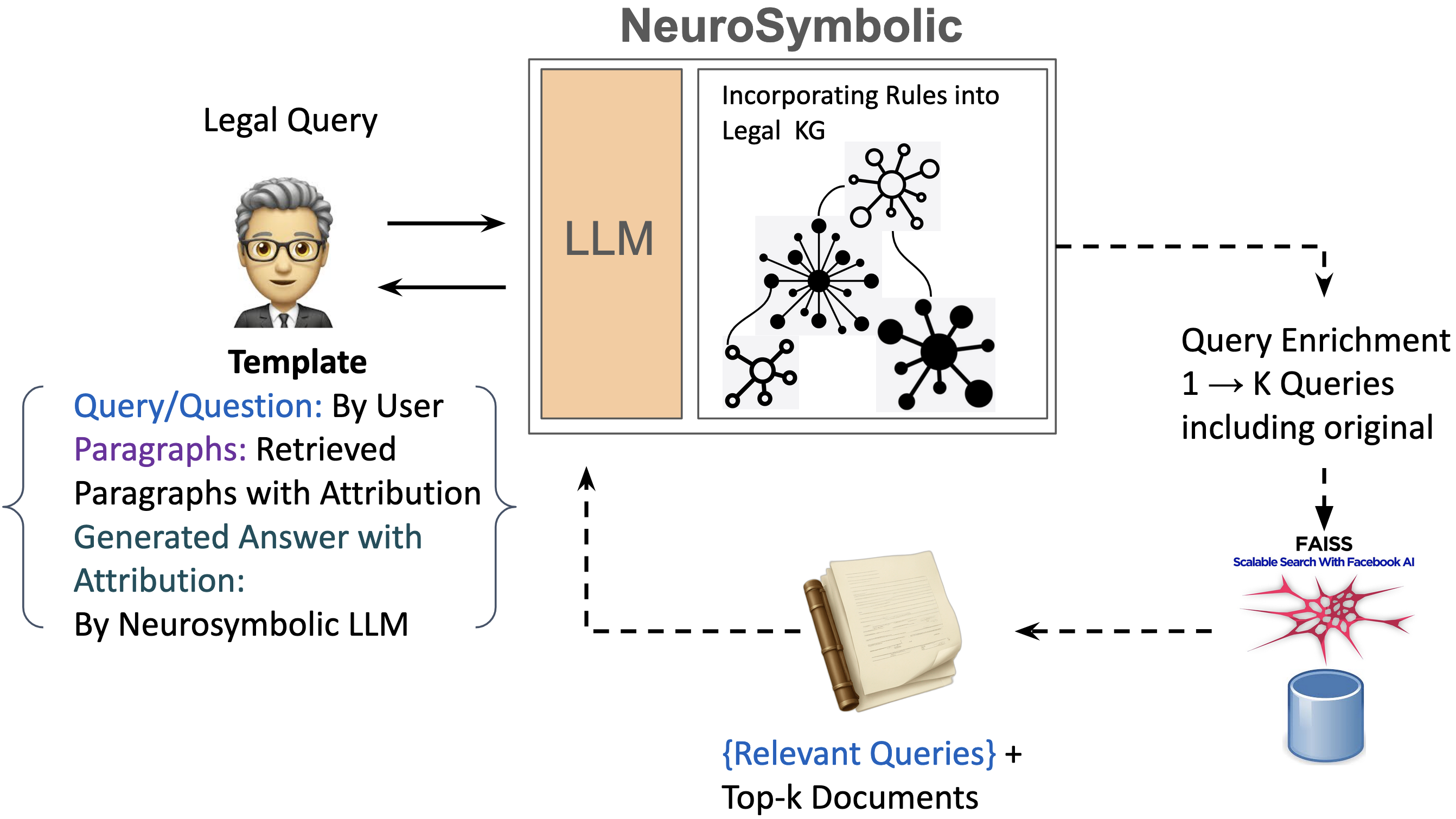}
    \caption{NeuroSymbolic RAG Pipeline integrating TRISM principles. The framework combines an LLM with a Legal Knowledge Graph (KG), leveraging query enrichment and retrieval techniques (e.g., FAISS \citep{douze2024faiss}) for scalable and secure document retrieval. It incorporates trust, responsibility, and security through enriched rule-based reasoning in the KG, ensuring robust query-answering in legal contexts.}
    \label{fig:RAG}
\end{figure}

Neurosymbolic RAG offers an innovative architecture that integrates neural retrieval mechanisms, typically powered by LLMs, with symbolic reasoning frameworks. This architecture, as shown in Figure \ref{fig:RAG}, is particularly well suited to the legal domain, where precision, compliance, and explainability are critical, yet the domain is characterized by complex, evolving jurisprudence and inherently ambiguous language. By combining the interpretive capabilities of neural models with the logical rigor of symbolic systems, neurosymbolic RAG facilitates the retrieval, reasoning, and generation of advanced legal knowledge for diverse legal tasks.

\textbf{Architecture:} The architecture of a neurosymbolic RAG system is centered on the interplay between its neural and symbolic components. Neural retrieval employs LLMs to process unstructured legal texts, such as case law, statutes, and procedural documentation, and extract semantically relevant information. These models, trained on vast legal corpora, leverage contextual embeddings to identify connections between concepts even when terminology varies. For instance, the system can match a query concerning “duty of care” with cases discussing “reasonable care” due to the neural model’s ability to generalize across related terms. This retrieval process ensures that the system captures the breadth of relevant information while accommodating the linguistic variability inherent in legal texts.

\textbf{Validation:} Once information is retrieved, the symbolic reasoning layer validates and refines the results using formal legal ontologies and rule-based logic. The symbolic component encodes domain-specific knowledge in structured formats, such as hierarchical taxonomies and procedural rules, enabling the system to enforce compliance with jurisdictional, temporal, and procedural constraints. For example, the symbolic reasoning module can filter retrieved case law to exclude precedents that fall outside the relevant jurisdiction or statutory timeframe, ensuring the validity of the results. Additionally, deductive reasoning techniques are applied to infer implicit relationships and validate logical consistency within the retrieved data.

\textbf{Generation:} The generation phase synthesizes outputs by combining the contextual richness of neural models with the structured reasoning of symbolic systems. Neural generation produces fluent and comprehensive responses, while symbolic reasoning constrains these outputs to align with legal norms and procedural requirements. This synthesis is particularly valuable for generating legal arguments, drafting procedural documents, or summarizing case law. For example, when drafting a legal brief, the system incorporates retrieved case law and procedural rules, presenting them in a manner that adheres to jurisdictional and argumentative standards. This dual-layered approach ensures that the outputs are both semantically rich and procedurally sound.

\subsection{Advantages of using Neurosymbolic RAG}

\paragraph{Capabilities}: One of the key strengths of neurosymbolic RAG in the legal domain is its ability to handle complex, multifaceted queries. Legal questions often require the synthesis of diverse information sources and the application of procedural constraints. For example, a query seeking precedents for admissible evidence in a specific jurisdiction involves retrieving case law, interpreting procedural statutes, and ensuring relevance to the query’s factual and temporal context. Neurosymbolic RAG addresses this by leveraging neural retrieval to access broad, semantically relevant data and symbolic reasoning to refine and contextualize the results. This ensures that the system delivers precise and actionable insights tailored to the user’s specific needs.

\paragraph{Adaptability}: The adaptability of neurosymbolic RAG is another critical advantage in the dynamic legal environment. Laws, precedents, and procedural standards evolve, and the system is designed to accommodate such changes. Neural models can be fine-tuned with updated corpora to reflect shifts in legal language and interpretive trends, while symbolic ontologies can be revised to incorporate new rules and statutory interpretations. This continuous updating process ensures that the system remains aligned with current legal standards, enabling practitioners to rely on it for accurate and timely guidance.

\paragraph{Transparency}: Transparency is a defining feature of neurosymbolic RAG, particularly in legal contexts where trust and accountability are paramount. Unlike purely neural systems, which often function as black boxes, the symbolic reasoning layer provides traceable and explainable decision-making paths. For example, when the system suggests a particular case as a precedent, it can articulate the reasoning process, citing the specific principles, jurisdictional rules, and procedural constraints that informed its decision. This transparency not only builds trust in the system’s recommendations but also ensures that legal professionals can confidently use its outputs in their work.

\section{Neurosymbolic AI for Legal Domain}

%\todo[inline]{Something about Reasons, ieee intelligent paper}

The application of Neurosymbolic AI in the legal domain introduces an innovative synthesis of neural learning and symbolic reasoning, essential for handling the intricate and interpretative nature of legal knowledge. Legal processes demand not just textual comprehension but also logical consistency, rule-based validation, and the ability to support nuanced decision-making under jurisdictional constraints ~\citep{wszalek2021cognitive, zhang2023application}. 
Neurosymbolic systems are exceptionally suited to address these challenges by combining the neural networks' ability to recognize patterns with the structured logical reasoning provided by symbolic frameworks.

Legal reasoning is uniquely challenging because it requires aligning textual and contextual information with codified rules and precedence hierarchies. The duality of neural and symbolic approaches is uniquely positioned to meet these demands. Neural models, such as LLMs, excel in processing and generating textual data, often capturing subtleties in legal language that would otherwise require significant human expertise. However, these models fall short in tasks requiring logical consistency and explainability, such as validating precedents or resolving jurisdictional conflicts. Symbolic systems, on the other hand, are inherently rule-based and operate within a framework of well-defined logic, making them ideal for encoding legal principles, procedural norms, and hierarchical relationships. Neurosymbolic AI synergizes these two approaches by utilizing the interpretative power of neural networks alongside the structured rigor of symbolic reasoning.

The potential of this integration can be fully realized through the incorporation of Knowledge Graphs (KGs), which act as structured repositories of legal knowledge. By organizing legal entities such as statutes, cases, and principles into graph-based relationships, KGs provide the foundational structure for reasoning. Retrieval-Augmented Generation (RAG) architectures further enhance this framework by leveraging the KG to ground generative tasks in precise, context-specific legal knowledge. For instance, when tasked with generating a legal brief, a Neurosymbolic AI system retrieves relevant precedents and statutes from the KG, synthesizing them into a coherent, procedurally sound argument. This process not only improves the accuracy of legal outputs but also ensures that every element is traceable to its source, a feature critical for maintaining trustworthiness in legal contexts.

A hallmark of legal reasoning is its reliance on precedential hierarchies and jurisdictional rules. Neurosymbolic AI addresses this by embedding legal ontologies within its symbolic reasoning layer. These ontologies encode the structural relationships and hierarchies among legal concepts, such as the interplay between state and federal laws or the precedence of constitutional rulings over statutory interpretations. This structured representation enables the system to reconcile conflicts, such as determining which law prevails in cases of jurisdictional overlap. For example, when analyzing a dispute involving both state and federal regulations, the system retrieves and ranks relevant statutes and cases. The neural component contextualizes the retrieved data, while the symbolic layer applies jurisdictional precedence rules, ensuring that the resulting analysis adheres to legal norms and standards.

Neurosymbolic AI’s ability to generate explainable outputs further underscores its relevance to the legal domain. Unlike traditional neural systems, which often operate as opaque black boxes, Neurosymbolic AI systems provide reasoning paths for their outputs. Each recommendation or decision, whether it involves citing a statute or resolving a conflict, is accompanied by a detailed explanation derived from the symbolic reasoning process and the KG. This transparency is particularly valuable in high-stakes applications, such as judicial decision-making or legislative drafting, where every assertion must withstand scrutiny. By enabling traceability and accountability, Neurosymbolic AI aligns itself with the stringent demands of the legal profession.

The significance of Neurosymbolic AI extends beyond practical applications to advancing legal domain. By automating the retrieval, reasoning, and synthesis of legal knowledge, it accelerates research and enables scholars to tackle complex legal questions with greater depth and efficiency. Its ability to model abstract legal concepts also paves the way for developing new legal theories and frameworks. For example, in analyzing the evolving concept of ``reasonable care” in tort law, the system could integrate historical precedents with contemporary interpretations, offering scholars a comprehensive perspective.

Drawing inspiration from REASONS ~\citep{tilwani2024reasons}, Neurosymbolic AI aligns with the benchmark’s emphasis on generating precise, context-aware, and explainable responses. Like REASONS, which integrates retrieval and reasoning to provide faithful and grounded outputs, Neurosymbolic AI ensures that every conclusion is anchored in validated knowledge. This alignment reinforces its suitability for domains where accuracy, interpretability, and compliance are paramount.

% summarization: 
\section{AI-TRISM with Neurosymbolic AI}
% classification: 

In order to use a neurosymbolic AI-based system, domain knowledge needs to be encoded in machine-processible form. One of the methods to explicitly encode the legal domain is to construct a legal KG.

\subsection{Building Dynamic Legal Knowledge: Graph Construction and Evolution}
KG is a graph with entities as nodes and relations between the entities as edges \citep{dong2014knowledge}. Compared to neural network learning patterns from the data, KG explicitly makes complex relationships among the entities, thereby offering a structured and semantically enriched framework to address these challenges. 

In the legal domain, a KG can significantly enhance decision-making, information retrieval, and case analysis with legal data in an interconnected framework. Legal KG provides relationships between the entities, such as cases, laws, legal principles, and stakeholders, to enable comprehensive analysis and support the roles of prosecutors, defenders, and judges, by representing legal entities (e.g., statutes, cases, legal principles) as nodes and their relationships as edges, KGs enable advanced reasoning and querying capabilities. We present a formal methodology for constructing KGs,  highlighting the integration of legal ontologies, graph construction techniques, and reasoning mechanisms to enhance legal research and support decision-making. 

\paragraph{Formal Framework for KG Construction}
A KG \( G \) is formally defined as a directed labeled graph:
\[
G = (V, E, \mathcal{L}),
\]
where: \( V \) is the set of nodes representing legal entities such as statutes, cases, or legal principles, \( E \subseteq V \times V \) is the set of relationships between nodes. \( \mathcal{L} \) is the set of labels that annotate the nodes and edges with semantic meaning. Each node \( v \in V \) is associated with an attribute set \( \mathcal{A}(v) \), containing metadata such as jurisdiction, date, and legal domain:

%\todo[inline]{Replace ''attribute`` with real things.}
\[
\small
\mathcal{A}(v) = \{\text{\texttt{Vedict: Defense Won}},
\]
\[
\small
\mathcal{\text{\texttt{Date: 11-Jan-2002}}, \dots,}
\]
\[
~\mathcal{\text{\texttt{Domain: Criminal Law}}\}}.
\]

A KG can be constructed using \texttt{Closed Information Extraction} (Closed IE) or \texttt{Open Information Extraction} (Open IE) (\cite{etzioni2008open}) based approaches. Closed Information Extraction assumes that the system has access to a predefined set of information to be extracted, such as the name of the defendant. In contrast, in an Open IE setup, the system does not have access to specific information to extract but instead aims to extract all possible information from the text.

\paragraph{Ontology Integration}
%\todo[inline]{Add an example that looks like legal thing}
For an information extraction system to extract information, a closed IE system needs access to an ontology. An ontology \( \mathcal{O} \) provides a schema for the KG, defining the structure and semantics of the graph:
\[
\mathcal{O} = (C, R, H, \mathcal{P}),
\]
%\todo[inline]{Deepa and Yash - make the wrap double quote words with texttt command. }
where \( C \) is the set of concepts (e.g., \texttt{"Statute", "Court Decision"}). \( R \) is the set of relations (e.g., \texttt{"cites", "interprets"}). \( H \subseteq C \times C \) is a hierarchical structure, where \( (c_1, c_2) \in H \) indicates \( c_2 \) is a subclass of \( c_1 \). \( \mathcal{P} \) is the set of properties for concepts, such as \texttt{"Effective Date"} or \texttt{"Jurisdiction".}

\paragraph{Formal Representation of Relationships}
In a KG each relationships between entities are modeled as triples:
\[
t = (s, p, o),
\]
where \( s \in V \): the subject (e.g., a case), \( p \in R \): the predicate (e.g., ``\texttt{cites}''), \( o \in V \): the object (e.g., a statute). For example, if a case \( c_1 \) \texttt{cites} a statute \( s_1 \), this relationship is represented as:
\[
(c_1, \texttt{cites}, s_1) \in E.
\]

\subsection{Graph Construction Methodology}

\begin{figure}[ht]
    \centering
    \includegraphics[width=\linewidth]{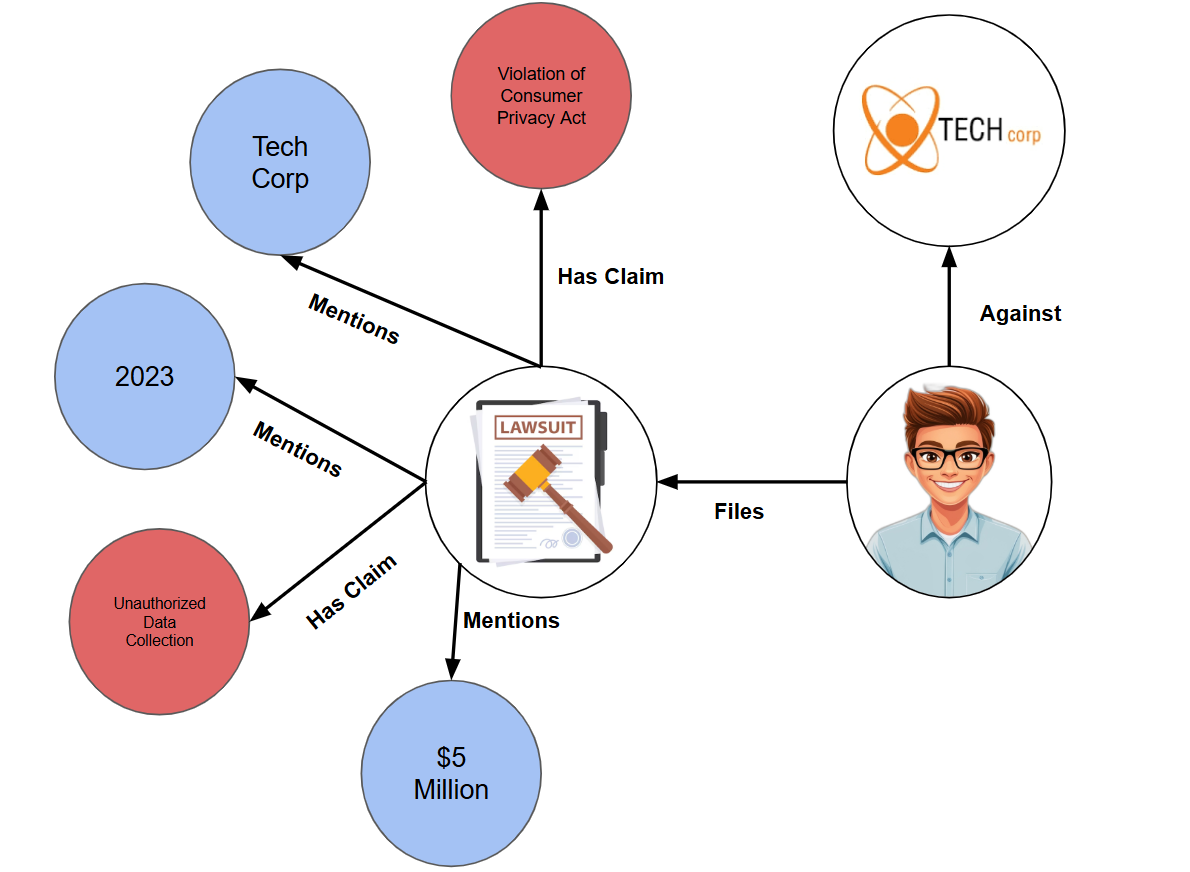}
    \caption{A representation of a Legal Knowledge Graph (KG) showcasing the relationships in a legal case, including entities such as a lawsuit, claims of violation of the Consumer Privacy Act and unauthorized data collection, the involved parties (an individual and Tech Corp), the year (2023), and the associated claim of \$5~\text{million} in damages.}
    \label{fig:example}
\end{figure}

\begin{algorithm}
\caption{KG Construction for Legal Domain}
\label{alg:1}
\begin{algorithmic}[1]
\State \textbf{Input:} Legal corpus $D = \{d_1, d_2, \dots, d_n\}$, Ontology $\mathcal{O} = (C, R, H, \mathcal{P})$
\State \textbf{Output:} KG $G = (V, E, \mathcal{L})$

\State \textbf{Step 1: Ontology Development}
\State Define concepts $C$, relationships $R$, hierarchy $H$, and properties $\mathcal{P}$.

\State \textbf{Step 2: Entity Extraction}
\For{each document $d_i \in D$}
    \State Extract entities $V = \{s_i, c_i, p_i\}$ using Named Entity Recognition (NER).
\EndFor

\State \textbf{Step 3: Relation Extraction}
\For{each sentence $S$ in $D$}
    \State Use dependency parsing and semantic role labeling to identify relations $E$.
    \State Map relations to triples $t = (s, p, o)$.
\EndFor

\State \textbf{Step 4: Ontology Alignment}
\For{each entity $v \in V$ and relation $e \in E$}
    \State Align $v$ and $e$ with concepts and relationships in $\mathcal{O}$.
\EndFor

\State \textbf{Step 5: Graph Augmentation}
\For{each node $v \in V$}
    \State Annotate with metadata $\mathcal{A}(v) = \{\text{Jurisdiction, Date, Domain}\}$.
\EndFor

\State \textbf{Step 6: Reasoning and Inference}
\For{each relation $(c_1, \text{cites}, s_1) \in E$}
    \If{$(s_1, \text{cites}, s_2) \in E$}
        \State Add $(c_1, \text{cites}, s_2)$ to $E$.
    \EndIf
\EndFor

\State \textbf{Step 7: Querying}
\State Provide SPARQL-like interface for querying $G$.

\State \textbf{Return} KG $G = (V, E, \mathcal{L})$

\end{algorithmic}
\end{algorithm}

An example of a legal KG is shown in Figure \ref{fig:example}. 
One of the necessary steps to create a KG is to extract entities (including events) from the text to represent nodes and relations as edges. Constructed KGs can be refined with Ontology alignment and graph augmentation and can be catered to the public with enabled reasoning and querying support. Each of the phases is described below:

\paragraph{Entity Extraction}: Given a corpus of legal documents \( D = \{d_1, d_2, \dots, d_n\} \), entities \( V \) are extracted using Named Entity Recognition (NER). For example, entities include:
\[
V = \{s_1, s_2, c_1, c_2, p_1, p_2\},
\]
where \( s_i \) are statutes, \( c_i \) are cases, and \( p_i \) are legal principles.

\paragraph{Relation Extraction}
Relations \( E \) are extracted using dependency parsing and semantic role labeling. For a sentence \( S \), dependency parsing identifies syntactic structures \( \mathcal{D}(S) \), while semantic role labeling assigns relationships. For example:
S: “Case $c_1$ cites Statute $s_1$”, yields: ($c_1$, \text{cites}, $s_1$).

\paragraph{Ontology Alignment}
Extracted entities and relations are aligned with the ontology \( \mathcal{O} \), ensuring semantic consistency. For example:
\[
c_1 \in C_{\text{Court Decision}}, \quad s_1 \in C_{\text{Statute}}.
\]

\paragraph{Graph Augmentation}
The graph \( G \) is enriched with metadata \( \mathcal{M} \). For example,a case \( c_1 \) is augmented as:
\[
\mathcal{M}(c_1) = \{\text{``Date: 2023-06-30,"} 
\]
\[
\mathcal{\text{``Jurisdiction: U.S. Supreme Court"}}\}.
\]

\paragraph{Reasoning and Querying}
Reasoning enables advanced inferences over the KG. Transitive reasoning applies to relations such as "cites":
\[
\text{If } (c_1, \text{cites}, s_1) \text{ and } (s_1, \text{cites}, s_2), \text{ then infer } (c_1, \text{cites}, s_2).
\]

Querying is facilitated using SPARQL-like expressions. For example, to find all statutes cited by a case:
\[
\text{MATCH } (c) -[:\text{cites}]-> (s) \text{ RETURN } s.
\]

Together, the algorithm for building KG is provided in the representation shown in Algorithm \ref{alg:1}

%\ref{fig:KG_Construction}.

\begin{figure*}[t]
    \centering
    \includegraphics[width=\linewidth]{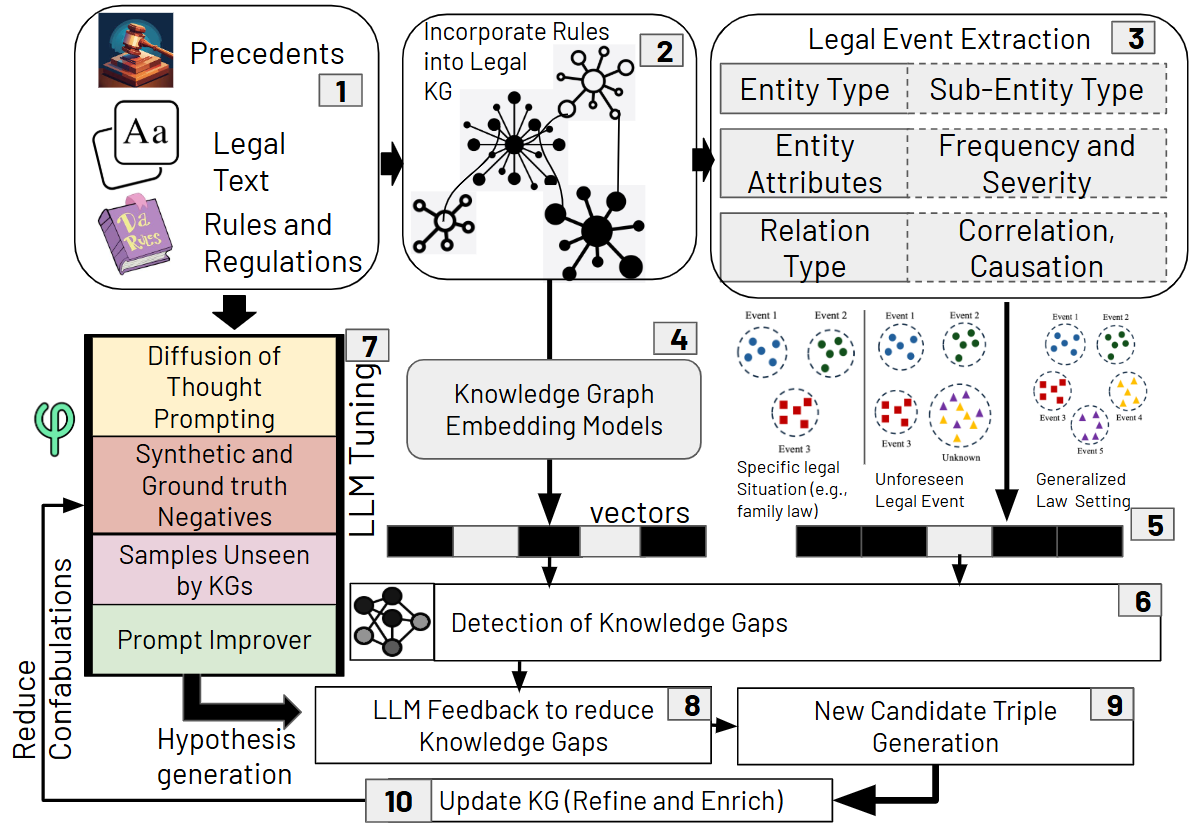}
    \caption{A 10-step process to automate the process of updating Legal KG using LLMs. The architecture highlights (a) the transformation of legal KG to embedding, (b) the identification of new legal events, (c) the detection of knowledge gaps, and (d) addressing the knowledge gaps with new candidate triples. A pre-trained LLM powers these major phases through three key mechanisms: (a) tuning with synthetic negative samples to establish clear decision boundaries and validate existing knowledge in the KG, (b) identification of unseen instances to explore and fill gaps in the current legal knowledge representation, and (c) dynamic prompt improvement to optimize hypothesis generation. The updated legal KG is then used to refine the LLM and even reduce hallucinations.}
    \label{fig:KG_Construction}
\end{figure*}

\subsection{Symbiosis of LLM and KG for Neurosymbolic RAG in Legal Domain}

One challenge with KGs is the need for their continuous updating to reflect new information and changing relationships within the domain they represent, as static knowledge quickly becomes outdated and less valuable. Manual updates to KGs are not feasible at scale, particularly within the legal domain, where legislative bodies and courts continuously enact new statutes, implement regulatory amendments, establish precedential decisions, and create exceptions to existing rules—all of which require thorough validation and careful integration into KG. 

KG scalability can be enhanced while maintaining its expressive capabilities by combining two key elements: (A) leveraging LLMs generative capabilities and extensive world knowledge, and (B) integrating this with KG through KG embedding techniques. The process of enriching KGs with information from LLMs is guided by a metric known as the Knowledge Gap. This gap represents areas where KG embedding methods can perform basic tasks like classification, recommendation, or attribute assignment, but do so with low confidence levels, indicating incomplete or uncertain knowledge in those areas.

The overall process of automatic updating of legal KG is shown as a 10-step process in Figure \ref{fig:KG_Construction}. 

\begin{description}
    \item[Step 1 \& Step 2] Legal KG construction: We have described the process for carrying out the Step 1 and 2 in Section 1.4.1. 
    \item[Step 3] Legal Event Extraction: This process involves instantiating the process of verb phrase extraction over the legal KG to extract triples containing information about a legal event.A legal event is manifested in the form of formal documentation (contracts, court orders, legislative acts), significant actions (violations, compliance, fulfillment of obligations), procedural events (filing of lawsuits, administrative decisions, regulatory approvals), or status changes (bankruptcy declarations, corporate formations, property transfers). Triples in KG that belong to such events are extracted and formed clusters. For clustering we recommend soft clustering techniques, such as locality sensitive hashing for creating event clusters. 
    \item[Step 4] Knowledge Graph Embedding Models: These are word embedding models trained over open source KGs. The three popular KG embedding models we recommend to use for creating embedding of legal KG is TransE (distance based, which can be modified to similarity based; ~\citep{bordes2013translating}), HolE ~\citep{nickel2016holographic}, and ANALOGY ~\citep{liu2017analogical}.
    \item[Step 5] Vectorization : A simple process of vectorizing the event clusters using KGE models and vectorizing legal KG models. 
    \item[Step 6] Detection of Knowledge Gaps in KGE models: The suggested Knowledge Graph Embedding (KGE) models are off the shelf and might not correctly represent rare or unseen legal events. To quantify the knowledge gaps in KGE, we compute divergence between knowledge representation of event clusters and their associated entity embeddings using multiple metrics. This divergence is calculated through a combination of (1) embedding space distance measures such as cosine similarity and Euclidean distance, (2) probability distribution divergence using Kullback-Leibler (KL) divergence or Jensen-Shannon divergence, and (3) structural entropy measures that capture the uncertainty in the graph topology. A high divergence score, coupled with low confidence in link prediction or entity classification tasks, indicates potential knowledge gaps. These gaps are particularly prevalent in areas where the KGE model encounters rare legal events, complex multi-hop relationships, or novel combinations of entities that weren't well-represented in the training data.
    \item[Step 7] Hypothesis generation: With the use of legal benchmarks, we fine-tune LLMs into generating hypotheses through a role-playing strategy, where LLMs are provided with cases to act on as defender and then subsequently as prosecutor. By understanding both sides of the coin in a legal argument, LLMs attain finesse in identifying gaps in knowledge in legal KG. 
    \item[Step 8 \& 9] Reduction of Knowledge Gaps and Identification of New Candidate Triples: The hypthosis generated by LLMs is carefully compared with the existing knowledge in the legal KG. This comparison process identifies three key aspects: exact matches (where the LLM's output perfectly aligns with existing knowledge), similarities (where there's partial overlap or close alignment), and differences (where the LLM suggests entirely new information). By evaluating these aspects and using specialized algorithms to predict possible connections in the legal KG, we can identify new, reliable pieces of information that could be added to enrich the KG while maintaining legal accuracy and relevance.
    \item[Step 10] Update KG: The process of updating the legal KG comprises two static functions: (a) Refine: If an existing triple is outdated or incorrect, a new candidate triple replaces such triples. This ensures the accuracy and currency of the KG by removing obsolete or erroneous information and replacing it with validated, up-to-date knowledge. (b) Enrich: This is the process of expanding the KG by adding new, validated triples that represent previously uncaptured legal relationships, events, or entities. This function helps fill knowledge gaps and increases the coverage of the legal domain without compromising the integrity of existing, accurate information. The enrichment process carefully validates new triples through confidence scoring and legal consistency checks before integration.
\end{description}

\noindent Figure \ref{fig:KG_Construction} present an illustration to achieve an up-to-date legal KG and a stable legal LLM, which can be used to power neurosymbolic RAG, shown in Figure \ref{fig:RAG}. 

\section{Related Work}
The use of LLMs in the legal domain is an active field of research. Even though LLMs are proficient in downstream NLP tasks, legal texts present subtle nuances and specific legal terms. These terms pose a challenge to LLMs when generating text for downstream tasks. This paper covers work related to the use and construction of KGs, work around legal citation and recommendation, and legal tasks that rely solely on neural-based approaches. Table \ref{tab:rw_table} show organization of the neural network approaches with and without using graph to encode knowledge. Because of the diversity of tasks in legal domain we limit ourself to the task listed hereby. However, this is not an exhaustive list of tasks and more information can be found in \citet{lai2024large}. 

\begin{table*}[t]
\centering
\begin{tabular}{@{}p{4cm}p{2cm}p{2cm}p{4cm}@{}}
\toprule
\textbf{Task} & \textbf{Category}  & \textbf{Knowledge Encoding} &  \textbf{Reference} \\
\midrule

\vspace{0.1cm}
\shortstack[l]{KG\\Construction}
   & Construction & -- & \citet{kondo2024collaborative}, \citet{hlyzov2017business}, \citet{filtz2017building}, \citet{tang2020salkg}, \citet{Song2020} \\
      \hline

      \multirow{2}{*}{\shortstack[l]{Charge/Judgement\\Prediction}} & \multirow{2}{*}{Classification}  &  Explicit &\citet{bi2024knowledge}, \citet{Gan2021}\\
      & &  Implict &\citet{He2023}, \citet{Wang2024}\\

{\shortstack[l]{Legal Statute\\Identification}}
         &  & Explicit & \citet{nguyen2023vn}\\

      \hline

       \multirow{2}{*}{Question Answer} & \multirow{2}{*}{\shortstack[l]{Question\\Answer}} & Explicit & \citet{Jiang2024}\\

        &  & Implicit &  \citet{Louis2024},  \citet{McElvain2019} \\
        
             \hline
\multirow{2}{*}{\shortstack[l]{Law Article/Case\\Retrieval}}   & \multirow{2}{*}{\shortstack[l]{Information\\Retreival}} & Explicit & \citet{Tang2024}\\

 & & Implicit &   \citet{Li2023}, \citet{deng2024learning}, \citet{Ma2023}\\

             \hline

\multirow{2}{*}{\shortstack[l]{Citation\\Recommendation}}  & \multirow{2}{*}{\shortstack[l]{Information\\Retreival}} & Explicit & \citet{dhani2021similar} \\
  &   & Implicit &\cite{huang2021context}, \citet{Wang2024}\\
            \hline

\multirow{2}{*}{\shortstack[l]{ Citation\\Interoperability}}
 &  & Explicit & \citet{sadeghian2018automatic}\\
 &  & Implicit &\citet{luo2023prototypebasedinterpretabilitylegalcitation}\\
             % \hline
% \shortstack[l]{Augment\\Dataset} & \shortstack[l]{ Data \\ Augmentation} & - &\citet{Kim2024}\\
% Patent Quality & & Explicit &\citet{wang2014exploring} \\

\bottomrule
\end{tabular}
\caption{Neural-based approaches, with and without the use of graphs, for various downstream legal tasks. Here, `-' indicates ``not applicable,'' ``Explicit`` denotes that knowledge is represented in the form of a graph, and ``Implicit'' signifies that knowledge encoded in a neural network is used, either by pre-training a language model on legal data or by relying on the knowledge present in an off-the-shelf language model.\label{tab:rw_table}}
\end{table*}

\subsection{KG Construction}
There are multiple techniques used to construct a KG in the legal domain, ranging from human-in-the-loop approaches to fully automated methods.

\citet{kondo2024collaborative} presents a human-in-the-loop approach for creating a legal KG by combining LLMs and legal experts to model complex relationships using a hierarchical tree structure. The hierarchical tree is constructed by aggregating legal information from the bottom up, with fine-tuned LLMs capturing the tree structure. Since the output may be imperfect, human feedback is employed to refine the final results and ensure accuracy.

In contrast, \citet{filtz2017building} adopts a more manual approach to building a KG schema. This iterative process begins with existing publicly available KGs, which are used for downstream tasks to evaluate quality. If the results are unsatisfactory, the KG is modified to meet specific requirements, and the process is repeated until satisfactory results are achieved.

\citet{Song2020} employs an automatic method to rapidly create a KG of legal entities, enabling tasks such as contract verification, contract generation, and contract analysis. This method uses a clustering-based approach applied to a predetermined set of contracts without requiring supplemental references and is robust against typos.

\citet{tang2020salkg} proposes a semi-automatic method leveraging NLP-based techniques to extract entities from text. These entities are then annotated by humans with specific relationships to generate final triples. The resulting graph is consumed using tools like Neo4j.

\subsection{Legal Classification}
The legal classification task aims to predict one or more labels from a set of predetermined labels.

\citet{bi2024knowledge} introduces a novel legal schematic knowledge-aware model designed for challenging legal cases, leveraging a transformer-based model and Graph Convolution Network. The authors create structural and semantic graphs using fact descriptions, integrating legal knowledge of groups, charges, and elements to derive final representations for classification.

\citet{Gan2021} employs declarative language to represent legal knowledge. Using first-order logic rules, \citet{Gan2021} captures interactions between different entities and integrates these rules into neural networks via co-attention mechanisms. By explicitly modeling interactions, the use of logic rules enhances the interpretability of the model.

\citet{He2023} models legal knowledge of charges as a set of tuples, where each charge is associated with its definition, subjectivity, subject, object, objectivity, and legal basis. This graph is encoded using a neural model and then integrated with fact descriptions to generate charge-level representations for classification.

\citet{Wang2024} introduces LegalReasoner, a system that combines LLMs with infused legal knowledge. The process involves pretraining an off-the-shelf model on a large legal corpus using a contrastive learning method. Subsequently, a graph neural network retrieves relevant cases, which are then provided as input to a transformer-based model and a generative network to deliver sound judgments with explanations.

\citet{nguyen2023vn} presents an approach aimed at assisting users with clarifications on Land Law matters. The authors construct a KG of questions and articles to capture intricate connections and associations among various legal entities and concepts, employing graph-based representation to determine the status of legal cases.

\subsection{Legal Question Answering}

Several approaches have been developed to perform question-answering tasks in legal contexts by combining information retrieval and language models.

\citet{Jiang2024} propose an approach to perform question answering on a legal knowledge base, where the question is a user query, and the answer is generated using a LLM with legal knowledge infused from a database of law articles. First, articles related to the query are retrieved and de-duplicated before generating the answer using the LLM.

Similarly, \citet{Louis2024} present a "retrieve-then-read" pipeline approach to support interpretability for legal documents written in French. For a given question, their approach first retrieves relevant articles using a bi-encoder model. Based on the retrieved documents, an answer is then generated.

Finally, and similar to the above approaches, \citet{McElvain2019} propose a method that uses an information retriever and natural language processing (NLP) models to obtain concise, single-sentence answers to basic non-factoid questions about the law.

\subsection{Legal Article and Case Retrieval}
Innovative approaches have been developed to enhance legal case retrieval and analysis by leveraging graph-based methods, pre-trained language models, and LLMs

\citet{Tang2024} propose a graph-based approach, CaseLink, to retrieve legal cases. They create a graph that connects query cases, candidate cases, and ground truth references between the query and candidates. During the testing phase, a given query case is converted into a representation and used to identify neighboring cases. To achieve this, \citet{Tang2024} employs a graph neural network to transform textual queries into a representable format.

\citet{Li2023} utilize pre-trained language models to retrieve legal cases. However, since off-the-shelf language models lack inherent legal knowledge, they incorporate domain-specific structures into the model. These structures are categorized as procedure, fact, reasoning, decision, and tail.  \citet{Ma2023} constructed a fine-grained corpus of articles from the original dataset to encode legal cases better. \citet{deng2024learning} prompt a LLM to extract crimes and corresponding law articles from legal cases. They leverage encoded knowledge from a legal expert database to map crimes accurately to the relevant law articles. Subsequently, the LLM is prompted with the extracted "crime-article" pairs to summarize the sub-facts of the crime from the legal case. The rationale behind using law articles is to provide high-level abstractions of criminal events, making it easier to identify critical sub-facts.

\subsection{Citation Recommendation and Interoperability}

\citet{dhani2021similar} use a KG to recommend similar cases. They first construct a KG from Indian court judgments and select features to train a graph neural network. The trained model generates representations of the judgments, which are then used to identify similar cases.

\citet{huang2021context} highlight that lawyers and judges spend significant time researching appropriate legal authorities to cite when drafting decisions. They propose a citation recommendation tool to improve the efficiency of opinion drafting. Four machine learning models are trained: a citation-list-based method (collaborative filtering) and three context-based methods (text similarity, BiLSTM, and RoBERTa classifiers). Their experiments demonstrate that leveraging local textual context improves recommendations and that deep neural models perform well. They observe that non-deep, text-based methods benefit from structured case metadata, while deep models only gain from such metadata when the context length is insufficient. Despite RoBERTa's pretraining advantages, it does not outperform a recurrent neural model after extensive training. A behavior analysis of the RoBERTa model reveals stable predictive performance across time and citation classes.

\citet{WangBAGJ24} address the limitations of collaborative filtering (CF)-based legal recommendation methods by introducing a model based on the Instruction GPT with Low-Rank Adaptation (IGPT-LoRA) architecture. This approach enhances legal citation recommendations by fine-tuning pre-trained language models (PLMs), improving performance while reducing computational demands. Additionally, \citet{WangBAGJ24} incorporate domain-specific instruction templates to guide the adaptation of PLMs.

% \citet{wang2024empowering}

\citet{luo2023prototypebasedinterpretabilitylegalcitation} address the legal citation prediction (LCP) requirements by designing the task to reflect the thought process of lawyers, incorporating both precedents and legislative provisions. They introduce a prototype-based architecture that enhances interpretability, delivering strong performance while aligning with the decision-making parameters used by lawyers. This approach refines citation predictions through feedback from legal experts.

\citet{sadeghian2018automatic} propose an algorithm that uses citing text along with its surrounding context to construct a citation graph from a text document. Once the graph is built, each edge is automatically labeled based on a predetermined set of labels.

\subsection{Other Related work}
Other distinctly related works include developing alert systems for users and assessing the quality of patents.

\citet{Hannah2024Prompt} propose a prompt engineering-based approach to generate alerts by combining the capabilities of LLMs with structured outputs. The system processes a user query to generate a response that identifies legal issues and retrieves relevant law citations for each issue. It then provides a concise, user-friendly summary of the applicable laws. These outputs populate an alert card, offering users a clear overview of legal concerns and corresponding references.

\citet{Wang2014Exploring} evaluates the value of patents using a probabilistic mixture approach that incorporates both technological and legal citations through citation network-based methods. This approach includes an iterative learning process that integrates a temporal decay function to enhance the analysis.

\section{Acknowledgement}
We acknowledge the support from UMBC Cybersecurity Leadership – Exploratory Grant Program. Any opinions, conclusions, or recommendations expressed in this material are those of the authors and do not necessarily reflect the views of UMBC.

\bibliographystyle{sageh}
\bibliography{reference}

\end{document}